\documentclass[runningheads]{llncs}

\usepackage[mobile]{eccv}

\usepackage{eccvabbrv}

\usepackage{graphicx}
\usepackage{booktabs}
\usepackage{multirow}
\usepackage{booktabs}
\usepackage{adjustbox}
\usepackage{array}
\usepackage{amssymb}
\usepackage{tabularx}
\usepackage{wrapfig}

\usepackage[accsupp]{axessibility}  

\usepackage[pagebackref,breaklinks,colorlinks,citecolor=eccvblue]{hyperref}

\usepackage{orcidlink}

\begin{document}

\title{Rethinking Pruning for Backdoor Mitigation: An Optimization Perspective} 

\titlerunning{}

\author{Nan Li\inst{1} \and
Haiyang Yu\inst{1}\and
Ping Yi\inst{2}}

\authorrunning{F.~Author et al.}
\institute{School of Cyber Science and Engineering, Shanghai Jiao Tong University, Shanghai 200240, China\\
\email{lyyqmjshu@sjtu.edu.cn}, 
\email{jhy549@sjtu.edu.cn}, 
\email{yiping@sjtu.edu.cn}}

\maketitle

\begin{abstract}
    Deep Neural Networks (DNNs) are known to be vulnerable to backdoor attacks, posing concerning threats to their reliable deployment. Recent research reveals that backdoors can be erased from infected DNNs by pruning a specific group of neurons, while how to effectively identify and remove these backdoor-associated neurons remains an open challenge. Most of the existing defense methods rely on defined rules and focus on neuron's local properties, ignoring the exploration and optimization of pruning policies. To address this gap, we propose an Optimized Neuron Pruning (ONP) method combined with Graph Neural Network (GNN) and Reinforcement Learning (RL) to repair backdoor models. Specifically, ONP first models the target DNN as graphs based on neuron connectivity, and then uses GNN-based RL agents to learn graph embeddings and find a suitable pruning policy. To the best of our knowledge, this is the first attempt to employ GNN and RL for optimizing pruning policies in the field of backdoor defense. Experiments show, with a small amount of clean data, ONP can effectively prune the backdoor neurons implanted by a set of backdoor attacks at the cost of negligible performance degradation, achieving a new state-of-the-art performance for backdoor mitigation.
\end{abstract}

\section{Introduction}

In recent years, Deep Neural Networks (DNNs) have demonstrated remarkable capabilities in solving real-world problems. However, the wide application of DNNs has raised concerns about their security and trustworthiness. Recent works have shown that DNNs are vulnerable to backdoor attacks\cite{gu_badnets_2017}, in which a malicious adversary injects specific triggers into the victim model through data poisoning, manipulating the training process, or directly modifying model parameters. The backdoored model performs well on clean samples but can be triggered into false predictions by the poisoned samples containing trigger patterns. As pre-trained weights and outsourced training are widely applied to cut computational costs for training DNNs, the backdoor attack is becoming an undeniable security issue. To address this issue, numerous methods have been proposed for detecting and mitigating backdoor attacks. Backdoor detection methods \cite{wang_neural_2019,liu2019abs,hu_trigger_2021} identify whether a model is backdoored or a dataset is poisoned, while backdoor mitigation methods \cite{liu_fine-pruning_2018,li_neural_2020,wu_adversarial_2021} eliminate the injected triggers from backdoored models. 

Our work focuses on the task of backdoor mitigation. Recent research \cite{wu_adversarial_2021,li_reconstructive_2023} has observed a subset of neurons contributing the most to backdoor behaviors in infected DNNs. By pruning these backdoor-associate neurons, the backdoor behavior of the infected model can be effectively mitigated. Mainstream approaches, as will be further introduced in Sect. \ref{rw}, concentrate on identifying these backdoor neurons using rule-based methods to obtain a clean model. However, the property of backdoor neurons varies across different attacks, models, and layers, motivating us to think about alternative approaches.

We investigate the distribution of backdoor and clean neurons, discovering that mitigating backdoor behavior often compromises clean accuracy, which inspires us to define backdoor mitigation as an optimization problem and introduce Reinforcement Learning (RL) to solve it. Moreover, we have observed that backdoor neurons and clean neurons tend to connect with neurons of the same type as themselves, motivating us to model the DNN as graphs and employ the Graph Neural Network (GNN) to leverage topological information within neuron connections. Building upon these insights, we propose \textit{Optimized Neuron Pruning} (ONP), the first optimization-based pruning method that combines GNN and RL to learn from neuron connections and optimize pruning policies for backdoor mitigation.

Our experiments demonstrate that ONP can defend against a variety of attacks and outperform current state-of-the-art methods, including ANP\cite{wu_adversarial_2021}, CLP\cite{zheng_data-free_2022}, and RNP\cite{li_reconstructive_2023} across different datasets, revealing the potential of optimization-based pruning methods in backdoor mitigation.

In summary, our main contributions are:
\begin{itemize}
\item We define pruning for backdoor mitigation as an optimization problem and develop a framework based on RL to optimize pruning policies for effective backdoor mitigation.
\item By investigating the distribution and connections of backdoor neurons, we develop a model-to-graph method for converting neuron connections into graphs to expose backdoor neurons. We further combine GNN and RL to conduct pruning on both the infected DNN and graphs.
\item We empirically show that ONP is competitive among existing pruning-based backdoor mitigation methods against a variety of backdoor attacks, which demonstrate the significance of optimization in enhancing backdoor mitigation performance.
\end{itemize}

\section{Related Work}\label{rw}
\subsection{Backdoor Attack}
Depending on trigger injection methods, backdoor attacks fall into two main categories: input-space attacks poisoning the training dataset and feature-space attacks manipulating the training process \cite{shafahi_poison_2018,cheng_deep_2021,zhao_defeat_2022} or directly modifying model parameters \cite{garg_can_2020,qi_towards_2022}. Input-space attacks can be further divided into static attacks using the same trigger for all samples and dynamic attacks using different triggers for different samples. Static triggers include patterns like black-white squares \cite{gu_badnets_2017}, Gaussian noise \cite{chen_targeted_2017}, adversarial perturbations \cite{turner_clean-label_2019}, or more complex patterns, while dynamic attacks such as the input-aware dynamic attack \cite{nguyen_input-aware_2020} and WaNet \cite{nguyen_wanet_2021} generate unique triggers for each input, making the defense more challenging.

\subsection{Backdoor Mitigation} 
Backdoor defense involves two primary tasks: backdoor detection and backdoor mitigation. Detection methods focuses on identifying backdoored models or poisoned datasets, while mitigation methods aims to remove the injected backdoor from the infected model with minimal degradation to its performance on clean samples.  Existing backdoor mitigation approaches include fine-tuning, pruning \cite{liu_fine-pruning_2018}, distillation \cite{li_neural_2020}, unlearning \cite{zeng_adversarial_2022} and training-time defenses \cite{li_anti-backdoor_2021,wang_training_2022,huang_backdoor_2022}. Recent works on pruning have demonstrated remarkable performance in backdoor mitigation. We divide these works into two basic categories: score-based methods and mask-based methods.

Score-based methods employ specific scores to measure the properties of backdoor neurons and determine the pruning policy based on each neuron's score. Fine-Pruning (FP) \cite{liu_fine-pruning_2018} uses neuron activation as the score and prunes dormant neurons to mitigate backdoor behavior. Neural Cleanse (NC) \cite{liu_fine-pruning_2018} synthesizes the backdoor trigger and prunes neurons activated by it. Entropy Pruning (EP) \cite{zheng_pre-activation_2022} identifies and prunes backdoor neurons based on the entropy of their pre-activation distributions. Channel Lipschitz Pruning (CLP) \cite{zheng_data-free_2022} introduces the channel lipschitz value to evaluate each neuron's sensitivity to input and prunes backdoor neurons with high sensitivity. Shapely Pruning \cite{guan_few-shot_2022} analyzes neuron's marginal contribution from a game-theory perspective. 

Mask-based methods create masks for each neuron, optimize the masks with a specific objective function, and prune neurons with low mask values to mitigate backdoor behavior. Adversarial Neuron Pruning (ANP) \cite{wu_adversarial_2021} uses masks to perturb neuron weights and prune neurons more sensitive to the perturbation. Reconstructive Neuron Pruning (RNP) \cite{li_reconstructive_2023} optimizes masks through an unlearning-recovering process to expose backdoor neurons.

In summary, score-based methods evaluate the computable properties of backdoor neurons, while mask-based methods optimize masks to capture complex properties. Although some advanced methods, like CLP \cite{zheng_data-free_2022}, ANP \cite{wu_adversarial_2021} and RNP \cite{li_reconstructive_2023} can effectively identify backdoor neurons and reduce the Attack Success Rate (ASR) to less than 1\%, the backdoor mitigation performance often comes at the cost of the Clean Accuracy (reducing more than 2\%), and hyperparameters need to be tuned to make these methods well-suited for different attacks. Both score-based and mask-based methods derive pruning policies with defined rules and are determined by the neuron's local property. In contrast, our ONP derives pruning policies through a try-and-learn process and can be considered an optimization-based method different from the rule-based methods mentioned above.

\subsection{Graph Neural Network}
GNN \cite{kipf_semi-supervised_2017} and its variants are widely used in processing graph structural data across diverse domains such as social networks, chemical molecules, and recommendation systems. GNNs extend the standard image convolution to graphs to aggregate neighbor structures and capture graph topology. Previous works \cite{yu_topology-aware_2022,jiang_channel_2022} in model compression have demonstrated that GNNs can effectively extract information from DNN structures, aiding in the optimization of pruning policies. Our proposed ONP uses the Graph Attention Network (GAT) \cite{velickovic_graph_2018}, an advanced GNN employing attention to better capture complex relationships between nodes, to extract information from neuron connections.

\section{Preliminaries}
\subsection{Notations}
Consider a $C$-class classification problem on a training set $\mathcal{D} = \{(\boldsymbol{x}_i, y_i)\}^N_{i=1}\subseteq \mathcal{X}\times\mathcal{Y}$, with $\mathcal{X}\subset \mathbb{R}^d$ as the sample space and $\mathcal{Y} \subset \{1, 2,...,C\}$ as the label space. Given a subset $\mathcal{D}_b\subseteq\mathcal{D}$, the standard poisoning-based backdoor attack involves injecting the trigger pattern into input samples with the poisoning function $\delta:\mathcal{X}\rightarrow \mathcal{X}$ and modifying corresponding labels with the label shifting function $S:\mathcal{Y}\rightarrow\mathcal{Y}$.

Let $F$ denote the victim model with parameter $\theta$. We assume $F$ as a convolutional network with $L$ layers, regarding the fully connected layer as the convolutional layer with $1\times 1$ kernels. Denote $f^{(l)}$ as the function of the $l^{th}$ convolutional layer and $\theta^{(l)}\in \mathbb{R}^{c\times c' \times h\times w}$ as the weight matrix of it, where $c$, $c'$, $h$, $w$ are the number of output and input channels, the height and width of the convolutional kernel, respectively. $\theta^{(l)}$ consists of $c$ filters $\{\theta^{(l)}_i\in \mathbb{R}^{ c' \times h\times w}\}_{i=1}^{c}$, and each filter consists of $c'$ kernels $\{\theta^{(l)}_{ij}\in \mathbb{R}^{h\times w}\}_{j=1}^{c'}$. The output of $f^{(l)}$ can be expressed as:
\begin{equation}
X^{(l)} = \phi\big(\underset{i=1} {\overset{c} {\Big \Vert}}\sum_{j=1}^{c'}(\theta^{(l)}_{ij}*X^{(l-1)}_j)\big),\label{eq1}
\end{equation}
where $*$ denotes the convolution operation, $\Vert$ denotes the concatenation operation, $\phi$ is the nonlinear activation function (e.g., Relu), and $X^{(l-1)}_j$ denotes the $j^{th}$ output channel of $f^{(l-1)}$ (also the $j^{th}$ input channel of $f^{(l)}$).

\subsection{Correlation between Backdoor and Clean Neurons}
Given a test set $\mathcal{D}_t$, we define the clean loss and the backdoor loss of $F$ as follows:
\begin{align}
\mathcal{L}_{cl}(F)&=\mathbb{E}_{(\boldsymbol{x}, y) \in \mathcal{D}_t}\mathcal{L}(F(\boldsymbol{x}; \theta), y),\\
\mathcal{L}_{bd}(F)&=\mathbb{E}_{(\boldsymbol{x}, y) \in \mathcal{D}_t}\mathcal{L}(F(\delta(\boldsymbol{x}); \theta), S(y))
\end{align}
To evaluate the impact of filters on clean accuracy and backdoor behavior, we consider pruning them and measuring the change in loss. Pruning the $i$-th filter of the $l$-th layer refers to setting $\theta^{(l)}_i$ to an all-zero matrix, thus removing the corresponding output feature map. We denote the pruned network by $F_{-(l,i)}$. For each filter, the \textit{Clean Loss Change} (CLC) and \textit{Backdoor Loss Change} (BLC) are defined as follows:

\begin{align}
\textrm{CLC}(F,l,i) &= \mathcal{L}_{cl}(F_{-(l,i)}) - \mathcal{L}_{cl}(F),\\
\textrm{BLC}(F,l,i) &= \mathcal{L}_{bd}(F_{-(l,i)}) - \mathcal{L}_{bd}(F),
\end{align}

\begin{figure}[t]
\centering
\includegraphics[scale = 1.2]{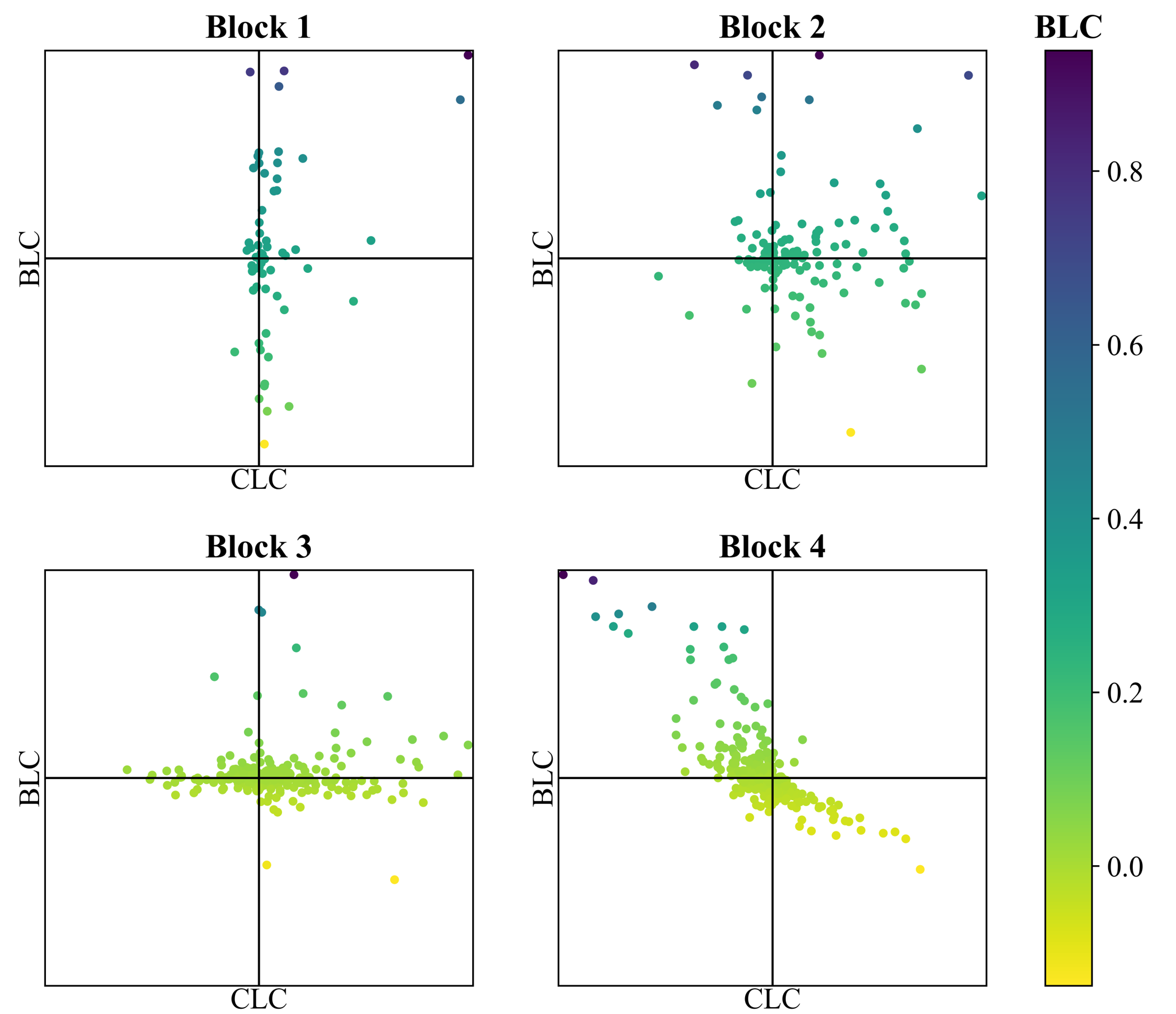}
\caption{BLC and CLC values of neurons in a backdoored ResNet18}
\label{fig2}
\end{figure}
Pruning neurons with positive BLC values mitigates the backdoor behavior by increasing backdoor loss, while pruning neurons with positive CLC values reduces clean accuracy. Note that BLC and CLC values cannot be directly used for backdoor mitigation because the impact of pruning multiple filters on loss is nonlinear and hard to compute. Fig \ref{fig2} shows the distribution of BLC and CLC values across 4 blocks of ResNet18. Neurons with positive BLC values fall into the first or second quadrant and can be potential backdoor neurons. Neurons with positive CLC values fall into the first or fourth quadrant. Neurons in the first quadrant can be considered both clean and backdoor neurons, making it more challenging to find a suitable pruning policy and motivating us to develop an optimization-based pruning method.

\subsection{Neuron connections reveals backdoor neurons}
$l_1$-norm is widely used to evaluate the importance of each filter in model compression methods \cite{li_pruning_2017}. Since convolution is considered a special linear transformation, we can use the $l_1$-norm of the convolutional kernel $\theta^{(l)}_{ij}$ to measure the correlation between $X^{(l-1)}_j$ and $X^{(l)}_i$. The $l_1$-norm of a convolutional kernel can be written as:
\begin{equation}
    \Vert \theta^{(l)}_{ij}\Vert_1 = \sqrt[1]{\sum_{n=1}^{c'} \sum_{k_1=1}^{h} \sum_{k_2=1}^{w}\left|\theta^{(l)}_{ij}(n, k_1, k_2)\right|},
\end{equation}
where $\theta^{(l)}_{ij}(n, k_1, k_2)$ denotes a single weight of $\theta^{(l)}_{ij}$. Generally, kernels with smaller $l_1$-norms produce lower activation values and have less numerical impact on the output feature map of the filter. Therefore, $\Vert \theta^{(l)}_{ij}\Vert_1$ indicates the neuron connection strengths between the $j^{th}$ channel of $f^{(l-1)}$ and the $i^{th}$ channel of $f^{(l)}$. By examining the connection strengths between consecutive convolutional layers, we have find that backdoor neurons tend to form strong connections with other backdoor neurons in the previous layer to amplify backdoor activation, while clean neurons show minimal connections with these backdoor neurons, as illustrated in Fig \ref{fig4:a}.

\begin{figure}[t]
\centering
\centering
\subfloat[\label{fig3:a}]{\includegraphics{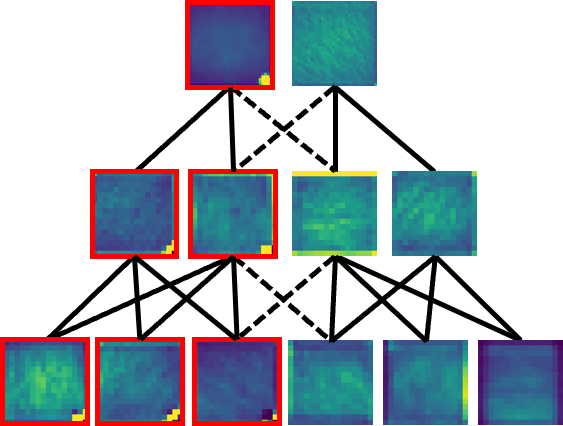}}
\subfloat[\label{fig3:b}]{\includegraphics[scale=1.0]{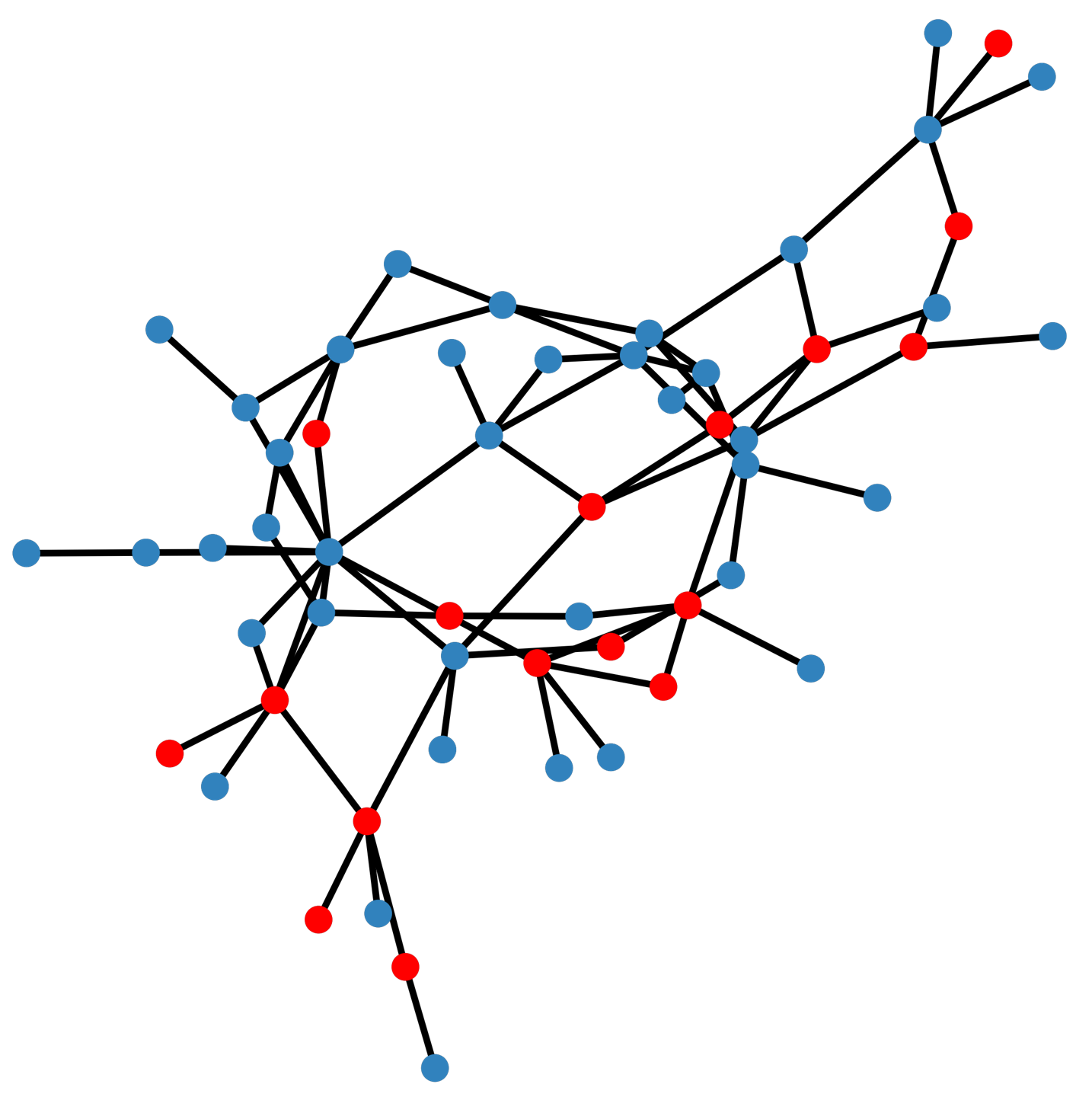}}
\caption{Neuron connections and an example of a constructed graph. (a)Backdoor and clean neurons primarily connect with neurons of the same type in the previous layer. (b)Part of the graph constructed for the second block of ResNet18, where red nodes represent potential backdoor neurons with large BLC values
}
\label{fig3}
\end{figure}

\begin{figure*}[t!]
\centering
\includegraphics[width = \textwidth]{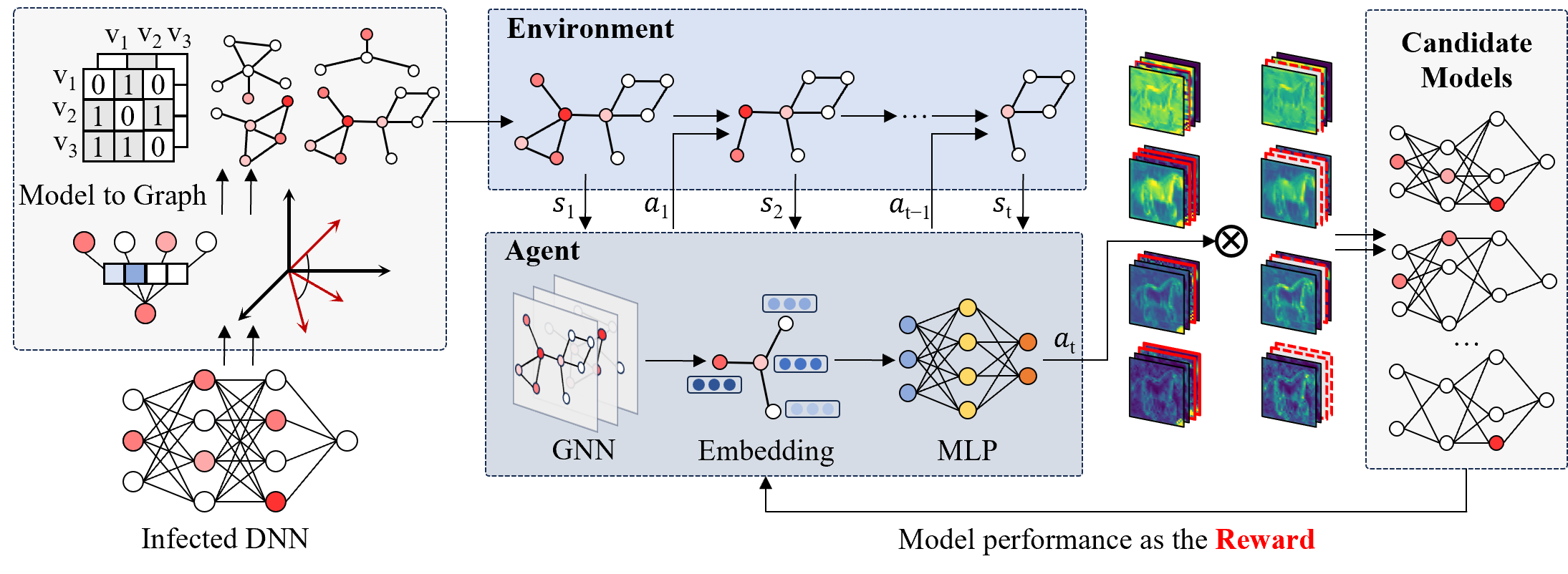}
\caption{Overview of our proposed ONP. ONP converts the infected model into graphs by neuron connections to exploit the inherent similarities among backdoor neurons, and then employs a RL agent containing GNN to learn from the graph and optimize the pruning policy}
\label{fig1}
\end{figure*}

\section{Methodology}
Our proposed ONP defense is illustrated in Fig \ref{fig1}. Unlike other rule-based backdoor defense methods, ONP views pruning for backdoor mitigation as an optimization problem rather than a classification problem. In other words, it does not identify backdoor neurons by their properties but employs RL agents to iteratively find a nearly optimal pruning policy.

ONP draws profound inspiration from successful concepts in model compression research, such as neuron similarity \cite{lei_compressing_2017,he_filter_2019} and the joint training of GNN and RL agent\cite{yu_auto_2021,jiang_channel_2022}. It first converts the infected DNN into multiple graphs to exploit the topological information within neuron connections, with each graph corresponding to a layer and each node corresponding to a specific channel. Then, ONP combines GNN and RL agents to learn pruning policies from the graphs. Each action taken by the agent results in channel-level pruning on the target DNN and changes to the graphs. The pruned DNN's performance on clean samples and backdoor samples is used as rewards for the agent's actions, encouraging the RL agent to continuously optimize the pruning policy for backdoor mitigation. In the following, we will explain further details on the graph construction and the RL agents.

\subsection{Defense setting. }
Following previous works \cite{wu_adversarial_2021,li_reconstructive_2023}, we assume the defender has downloaded a backdoored model from an untrustworthy third party without knowledge of the attack or training data. We assume a small amount of clean data is available for defense. The goal of backdoor mitigation is to remove the backdoor behavior from the infected model with minimum degradation to its clean accuracy.

\subsection{Graph Construction}\label{gc}
Previous research \cite{yu_auto_2021,yu_topology-aware_2022} on model compression has highlighted the value of topological information within DNNs for model pruning, inspiring us to explore whether connections between neurons can potentially expose backdoor neurons. To leverage the topological information, we develop a similarity-based method \cite{lei_compressing_2017,he_filter_2019} to incorporate neuron connection strengths into graphs. Consider a graph $\mathcal{G}=(\mathcal{V},\mathcal{E})$ constructed for the $l^{th}$ layer $f^{(l)}$, where the node set $\mathcal{V}=\{v_1, v_2,...,v_c\}$ corresponds to the filters of $f^{(l)}$. For the $i^{th}$ channel, we compute the connection strength $\boldsymbol q_i =(\vert\theta_{i1}\vert, \vert\theta_{i2}\vert,...,\vert\theta_{ic'}\vert)$, where $\vert\theta_{ij}\vert$ represents the l1-norm of $\theta_{ij}$. Given two certain thresholds $\epsilon$ and $\delta$, the edge set $\mathcal{E}$ can be determined by the cosine similarity of neuron connections:
\begin{equation}
\mathcal{E} = \left\{(v_i,v_j)\left|v_i,v_j\in\mathcal{V},\tfrac{\boldsymbol {q}_i \cdot \boldsymbol {q}_j}{\Vert\boldsymbol q_i\Vert\Vert\boldsymbol q_j\Vert}>\epsilon,\Vert\boldsymbol q_i\Vert_\infty>\delta\right.\right\},
\end{equation}
where $\Vert\boldsymbol q_i\Vert$ and $\Vert\boldsymbol q_i\Vert_\infty$ denote the l2-norm and the infinite norm of $\boldsymbol q_i$, respectively. Therefore, the connection strengths between neurons are transformed into edges in $\mathcal{G}$. Note that $\epsilon$ determines the number of edges and $\delta$ is used to filter dormant neurons. Part of a constructed graph is shown in Fig \ref{fig3:b} as an example, where potential backdoor nodes (neurons) are closely connected or neighboring nodes, indicating the effectiveness of our graph construction method. 

Following previous work on model compression \cite{jiang_channel_2022}, we consider the distribution of model activation as the node feature, which reflects the importance of channels. Feeding $n$ images traversing the model, the average activation of the $i^{th}$ channel forms a set $\mathcal{M}_i=\{m\}^n$. Splitting the sampling interval $[0, u]$ into $s$ sub-intervals, the distribution of $\mathcal{M}_i$ as well as the feature of the corresponding node in $\mathcal{V}$ can be represented by a vector $\boldsymbol p_i = (p_{i1}, p_{i2},..., p_{is})$, where $p_{ij}$ refers to:
\begin{equation}
p_{ij} = \frac{1}{n}\left|\left\{m \left| m\in \mathcal{M}_i, (j-1)\frac{u}{s}<m<j\frac{u}{s}\right.\right\}\right|
\end{equation}
\subsection{Model Pruning with Reinforcement Learning}
We employ GNN together with RL to learn from the constructed graph and find a suitable model pruning policy. In the following, we will explain the details of the RL agent.
\subsubsection{Overview} ONP conducts pruning on the infected model in a layer-wise way, with each agent corresponding to a specific layer and determining the filters to be pruned. The pruning process starts from the last convolutional layer of the model, and the agents corresponding to the shallower layers work on the network pruned by the previous agents. Each agent contains a GAT \cite{velickovic_graph_2018} and is optimized by the Proximal Policy Optimization (PPO) \cite{schulman_proximal_2017} algorithm, since GAT is a spatial GNN well-suited for dynamic graphs, and PPO demonstrates faster convergence and superior performance on our task compared to other RL methods. Generally, We recommend applying ONP to a subset of deeper layers (e.g., the last two blocks for ResNet-18) to balance backdoor erasing performance and clean accuracy, as will be further discussed in Sect \ref{ex}.
\subsubsection{Environment states} We use the graph established in Section 2 as the environment for the agent. The agent's action results in not only the pruning of specific neurons but also changes in the state of the environment. Once a neuron is pruned, we remove the corresponding node and edges from the constructed graph. As a result, the node embeddings learned by the GNN change accordingly to keep the RL agent informed of the current state of the DNN.
\subsubsection{Action Space} The actions taken by the RL agent are directly the indices of neurons to be pruned in each step within a discrete space. The actor network contains a GAT encoding environment states into node embeddings and a multi-layer perception neural network (MLP) projecting node embeddings into logits for each neuron, followed by a softmax to convert the logits into probabilities. The behavior function is a categorical distribution determined by the probabilities:
\begin{equation}
\pi(a|s) = \textrm{Categorical}(\textrm{Softmax}(\textrm{MLP}(\textrm{GAT}(\mathcal{G}))))
\end{equation}
To accelerate the convergence of the policy, we sample $K$ times from $\pi(a|s)$ to get the action $a \in \{1, 2, ...,  c'\}^K$ which determines the $K$ neurons to be pruned in each step.

\subsubsection{Reward}
To mitigate the infected model's backdoor behavior while preserving its performance on clean task as much as possible, we design the reward as a function of BLC and CLC computed on a limited amount of defense data:
\begin{equation}
Reward = \textrm{exp}(-\textrm{CLC}(F_{-A_{t-1}}, a_t)) - \textrm{exp}(-\lambda\textrm{BLC}(F_{-A_{t-1}}, a_t)),
\end{equation}
where $A_{t-1}$ refers to $\cup_{i=1}^{t-1}a_i$, $F_{A_{t-1}}$ denotes the partially-pruned DNN in $t-1$ step, and $\lambda$ is a manually defined coefficient. The exponential function is applied to both the CLC and BLC terms to maximize the penalty for degradation on clean performance and counteract the exponential growth of BLC as the number of pruned neurons increases. 

To compute BLC, we consider reverse engineering the backdoor triggers to get the poison dataset. Following previous work \cite{guan_few-shot_2022}, we use Neural Cleanse \cite{wang_neural_2019} for backdoor trigger synthesis on our defense data and choose the trigger with the smallest l1-norm for defense and the corresponding class as the attack label. Note that our method is also compatible with other advanced trigger synthesis methods.

\subsection{Pruning Strategy for ResNet}
\subsubsection{Residual Connections} Residual connections are widely used in many network structures and bring channel relevance between layers. Research in model compression \cite{li_pruning_2017,chen_only_2021} has extensively studied pruning strategies for residual networks, while the residual connection's impact on backdoor behavior is still unexplored. To address this gap, we have conducted a simple investigation. As is shown in Fig \ref{fig4:a}, the last two residual blocks of an infected ResNet-18 model are activated by the backdoor trigger in a similar way, and the indices of the activated channels are almost the same, which proves the channel relevance exists in deep layers and holds for backdoor activation.
\subsubsection{Accelerating ONP with Group-based Pruning} 
\begin{figure}[tb]
\centering
\subfloat[\label{fig4:a}]{\hspace{-6mm}\includegraphics[scale=0.8]{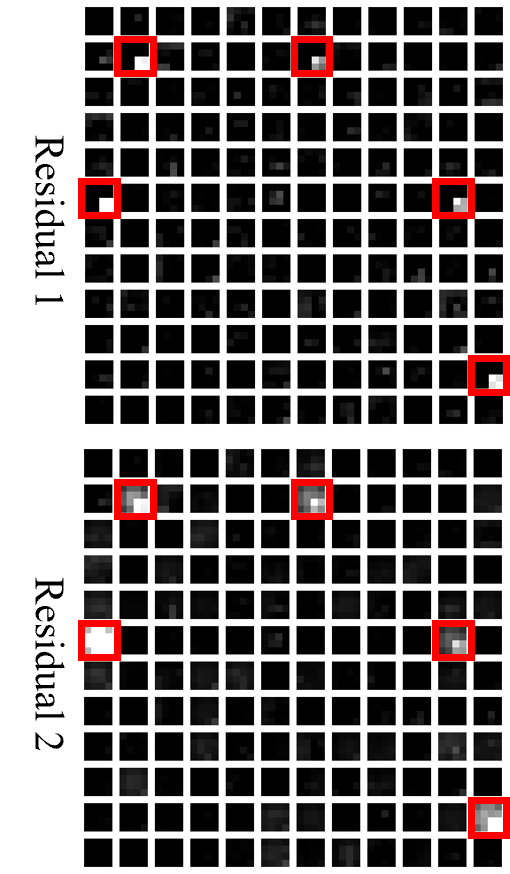}}
\subfloat[\label{fig4:b}]{\includegraphics[scale=0.8]{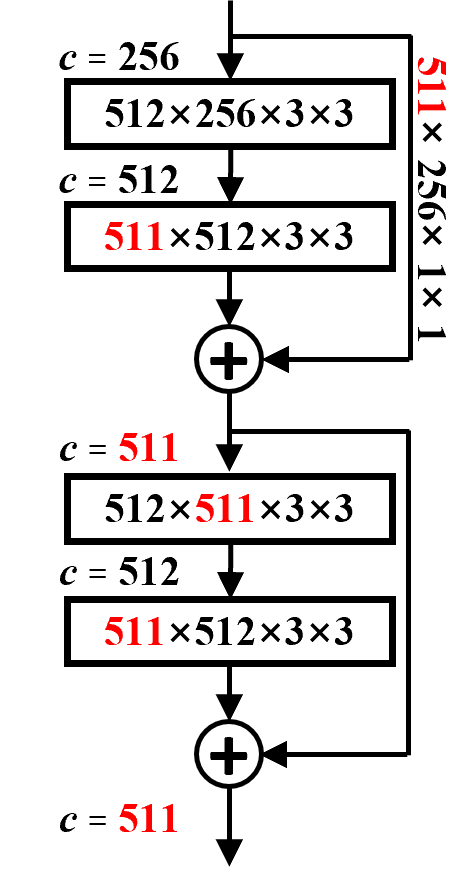}}
\caption{Channel relevance due to residual connections and the corresponding pruning strategy. (a)Activation map of the last two residual blocks in an infected ResNet 18 with poisoned samples as input, only 144 neurons are selected for simplicity (b)Group-based pruning strategy for the last two residual block in ResNet 18}
\label{fig4}
\end{figure}

Following previous work\cite{chen_only_2021} on model compression, we divide all layers into groups. The convolution layers within each group share the same number of output channels and are interconnected through residual connections. For ResNet, a group corresponds to a ResNet block comprising more than two residual blocks. Pruning an output channel of the whole group refers to pruning all related neurons across different layers, as is shown in Fig \ref{fig4:b}. For the model-to-graph method introduced in Sect \ref{gc}, we use the weight matrix of the last convolutional layer to compute neuron connection strengths and construct only one graph for each group. In this way, only 1 agent is needed for a ResNet18 Block of 4 convolutional layers, cutting 75\% computing costs.

\begin{table*}[tb]
\caption{Performances of 5 backdoor defense methods against 6 backdoor attacks. The experiments were performed on CIFAR-10 and Tiny ImageNet with ResNet-18 using only 1\% clean defense data. All these methods except FT do not need fine-tuning after the pruning. Note that ONP/o represents ONP with original trigger. ASR: attack success rate (\%); CA: clean accuracy (\%). The best results are \textbf{boldfaced} and the second-best results are \textit{\textbf{italicized}}}
\renewcommand{\arraystretch}{1.2} 
\centering
\begin{adjustbox}{width=0.98\linewidth}
\begin{tabular}{c|c|cc|cc|cc|cc|cc|cc|cc}
\toprule
\multirow{2}{*}{Datasets} & \multirow{2}{*}{\begin{tabular}[c]{@{}c@{}}Backdoor \\ Attacks\end{tabular}} & \multicolumn{2}{c|}{Backdoored} & \multicolumn{2}{c|}{FP} & \multicolumn{2}{c|}{ANP} & \multicolumn{2}{c|}{CLP} & \multicolumn{2}{c|}{RNP} & \multicolumn{2}{c|}{ONP (Ours)} & \multicolumn{2}{c}{ONP/o (Ours)} \\ \cline{3-16} 
 &  & ASR & CA & ASR & CA & ASR & CA & ASR & CA & ASR & CA & ASR & CA & ASR & CA\\ \midrule
\multirow{13}{*}[8ex]{CIFAR-10}
 & BadNets & 100.00 & 92.43 & 25.31 & 81.45 & 1.12 & 90.27 & 1.34 & 90.54 & 0.53 & 91.09 & \textbf{0.44} & \textit{\textbf{92.00}}  & \textit{\textbf{0.76}} & \textbf{92.20} \\
 & Trojan & 100.00 & 92.68 & 62.57 & 82.63 & 1.31 & 90.78 & 2.87 & 91.16 & 2.03 & 91.85 & \textbf{0.72}  &  \textbf{92.26} & \textbf{0.72} & \textbf{92.26}\\
 & Blend & 99.99 & 92.15 & 76.44 & 83.26 & 0.90 & 90.82 & 1.67 & 91.61 & \textbf{\textit{0.54}} & 91.53 & 1.58  & \textbf{91.89}  & \textbf{0.44} & \textit{\textbf{91.54}}\\
 & CL & 98.93 & 91.55 & 36.42 & 81.38 & 5.47 & 89.96 & 1.54 & 89.51 & \textit{\textbf{0.39}} & \textit{\textbf{90.63}} & 0.92 & 90.50  & \textbf{0.33} & \textbf{90.90}\\
 & Dynamic & 99.66 & 94.60 & 38.95 & 85.42 & \textbf{\textit{1.52}} & 93.67 & 4.71 & 93.20 & 2.34 & \textbf{94.19} & 2.16 & 93.47 & \textbf{0.35} & \textbf{\textit{93.98}} \\
 & WaNet & 98.81 & 92.12 & 69.74 & 80.53 & 4.79 & 91.06 & 4.14 & 90.58 & 3.02 & \textit{\textbf{91.60}} & \textbf{\textit{1.76}} & 91.34 & \textbf{0.88} & \textbf{91.75} \\\cline{2-16} 
 & Average & 99.57 & 92.59 & 60.98 & 82.45 & 2.52 & 91.09 & 2.71 & 91.10 & 1.48 & 91.82 & \textit{\textbf{1.26}} & \textit{\textbf{91.91}} & \textbf{0.58} & \textbf{92.11} \\ \hline
\multirow{13}{*}[8ex]{Tiny ImageNet} 
 & BadNets & 99.80 & 58.78 & 50.32 & 48.56 & 2.78 & 58.20 & 1.51 & 57.33 & \textbf{0.10} & \textit{\textbf{58.26}} & 0.96 & 57.84  & \textit{\textbf{0.53}} & \textbf{58.42} \\
 & Trojan & 99.99 & 59.13 & 93.44 & 49.75 & 8.49 & 57.46 & 1.29 & 58.35 & 0.60 & \textbf{58.84} & \textit{\textbf{0.46}} & 57.85 & \textbf{0.31} & \textit{\textbf{58.51}}\\
 & Blend & 99.99 & 57.12 & 83.71 & 51.24 & 0.84 & 56.03 & \textit{\textbf{0.77}} & 56.18 & 0.92 & 56.25 & 2.32  &  \textit{\textbf{56.34}}  & \textbf{0.55} & \textbf{56.57}\\
 & CL & 72.08 & 60.42 & 49.51 & 53.22 & 10.59 & 58.93 & 2.16 & \textbf{60.21} & \textbf{0.34} & 60.07 &  0.74 & 58.78 & \textit{\textbf{0.56}} & \textit{\textbf{60.19}}\\
 & Dynamic & 98.78 & 61.20 & 87.31 & 55.24 & \textit{\textbf{5.68}} & 60.07 & 10.45 & 58.86 & 6.33 & \textit{\textbf{60.45}} & 7.15 & 59.66 & \textbf{3.39} & \textbf{60.71} \\
 & WaNet & 97.53 & 59.56 & 68.72 & 55.23 & 13.54 & 57.38 & 3.76 & \textit{\textbf{58.43}} & 7.64 & \textbf{58.69} & \textit{\textbf{2.52}} & 55.83 & \textbf{1.41} & 56.39 \\\cline{2-16} 
 & Average & 94.70 & 59.37 & 72.17 & 52.21 & 6.94 & 58.01 & 3.32 & 58.23 & 2.66 & \textbf{58.76} & \textit{\textbf{2.06}} & 57.72 & \textbf{1.13} & \textit{\textbf{58.47}} \\ \bottomrule
\end{tabular}
\end{adjustbox}
\label{tab:2}
\end{table*}

\section{Experiments}\label{ex}
\subsection{Experimental Setup}
\subsubsection{Attack Setup.}We evaluate ONP against 6 famous attacks. These include 4 static attacks: BadNets \cite{gu_badnets_2017}, Trojan \cite{liu_trojaning_2018}, Blend \cite{chen_targeted_2017}, and Clean Label \cite{turner_clean-label_2019}, as well as 2 dynamic attacks: Dynamic \cite{nguyen_input-aware_2020} and WaNet \cite{nguyen_wanet_2021}. Default settings from original papers and open-source codes are followed for most attacks, including backdoor trigger pattern and size. The backdoor label of all attacks is set to class 0, with a default poisoning rate of 10\%. Attacks are performed on CIFAR-10 and Tiny ImageNet using ResNet-18. For training setups, Stochastic Gradient Descent (SGD) is utilized with an initial learning rate 0.1, weight decay 5e-4, momentum 0.9, batch size 128 for 200 epochs on CIFAR-10, and batch size 64 for 150 epochs on Tiny ImageNet. A cosine scheduler is employed to adjust the learning rate.
\subsubsection{Defense Setup.}
We compare ONP with 4 pruning-based backdoor mitigation methods, including Fine-Pruning \cite{liu_fine-pruning_2018}, and state-of-the-art methods ANP \cite{wu_adversarial_2021}, CLP \cite{zheng_data-free_2022} and RNP \cite{li_reconstructive_2023}. CLP is data-free and all other defenses share limited access to only 1\% clean samples from the benign training data. We adapt hyperparameters for these defenses based on open-source codes to obtain best performance against different attacks. For ONP, we set hyperparameter $\lambda$ to 5 for CIFAR-10 and 10 for Tiny ImageNet. $\epsilon$ and $\delta$ are adaptive adjusted to conserve 5\% edges 50\% nodes in each graph. The impact of these hyperparameters will be discussed in Sect. \ref{para}. ONP is applied to the last two blocks of ResNet18 for most attacks to balance backdoor elimination and clean accuracy. For Blend and WaNet, ONP is applied to all blocks, because their trigger patterns covers the whole image, thus more backdoor neurons are implanted in shallow layers. Policy of each agent is updated every 16 search episodes, and is optimized for 100 search episodes in total. Other settings follow the open-source code of PPO algorithm \cite{schulman_proximal_2017}. Backdoor trigger synthesis follows the original setup in Neural Cleanse paper\cite{wang_neural_2019}, optimizing triggers for each class and choosing the trigger with smallest $l_1$-norm for defense and assigning the corresponding class as the attack target label.
\subsubsection{Evaluation Metric.}
We adopt two metrics for evaluating backdoor mitigation performance: 1) Clean Accuracy (CA), which is the model's accuracy on clean test data; 2) Attack Success Rate (ASR), which is the model's accuracy on backdoored test data.
\subsection{Main Defense Results}
\subsubsection{Results on CIFAR-10} Table \ref{tab:2} presents the defense performance of 5 pruning methods against 6 backdoor attacks on CIFAR-10 and Tiny ImageNet. On CIFAR10, the standard ONP method outperforms other defense methods by cutting the average ASR down to 1.26\% with a slight drop on CA (lower than 1\% on average). In comparison, FP, ANP, CLP, RNP reduce the average ASR to 60.98\%, 2.52\%, 2.71\%, and 1.48\%, respectively. ONP shows weakness against Blend, CL, and Dynamic attacks due to its reliance on the backdoor trigger synthesized by NC. The ONP/o results clearly show that ONP can be further improved by applying advanced trigger synthesis methods. Despite the limitations of trigger synthesis methods, ONP still effectively defends against static and dynamic backdoor attacks, revealing the potential of optimization-based pruning in backdoor mitigation. Among other defense methods, RNP demonstrates promising and remarkable performance in defending against most attacks, while ANP and CLP exhibit their own strengths. However, FP shows the poorest performance in most settings, indicating that pruning dormant neurons may not be an effective choice for backdoor mitigation against advanced attacks. 
\subsubsection{Results on Tiny ImageNet}
Trigger reverse synthesis becomes more challenging on Tiny ImageNet, which further limits the performance of ONP. Despite these challenges, ONP remain competitive among state-of-the-art backdoor mitigation methods, with an average ASR reduction of 92.64\% and a acceptable average CA decrease of 1.65\%. RNP is also highlighted as an effective defense on Tiny ImageNet. 
\begin{wrapfigure}{r}{0.5\textwidth}
  \vspace{-10pt}
  \centering
  \includegraphics[width=\linewidth]{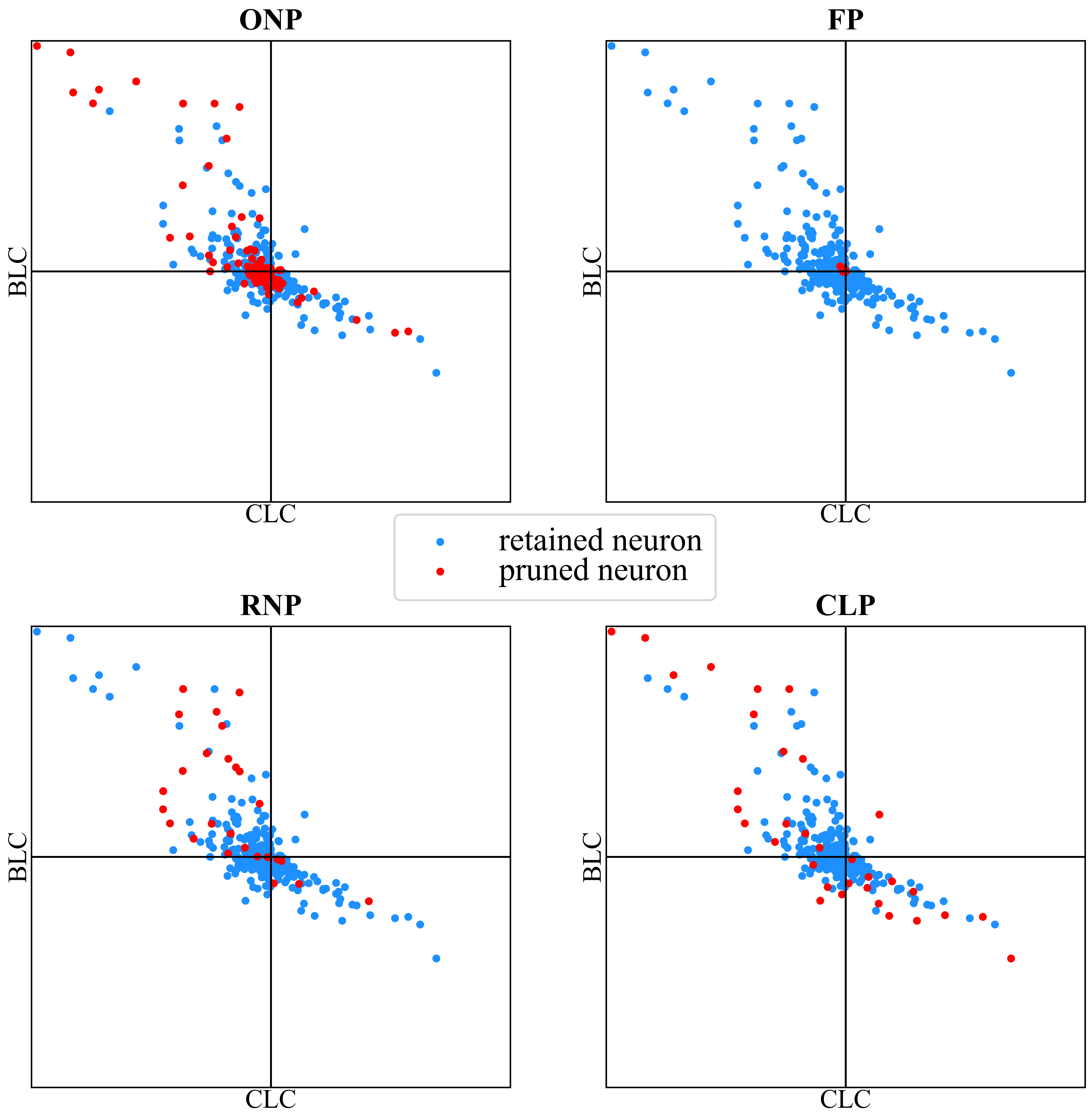}
  \caption{Pruning policies for the last convolutional layer of a backdoored ResNet18, derived from 4 different defense methods}
  \label{fig5}
  \vspace{-30pt}
\end{wrapfigure}
However, ONP/o achieves lower ASR than RNP under most settings, especially against dynamic attacks (\textless 5\% on average), which further emphasizes the effectiveness of the optimization of pruning policy.


\subsubsection{Analysis of Pruning Policies} We compare the pruning policies derived from different defense methods, as shown in Fig \ref{fig5}. For the last convolutional layer of the backdoored model, FP prunes 88 dormant neurons but does not hit any potential backdoor neurons, which is not sufficient for backdoor mitigation. RNP and CLP both prune 33 neurons and prove their effectiveness in identifying potential backdoor neurons. However, RNP doesn't find neurons with the biggest BLC values in the second quadrant, while CLP mistakenly prunes some clean neurons in the fourth quadrant. Our ONP prunes 136 neurons, including neurons with largest BLC values for erasing the backdoor and neurons with negative CLC values for compensating for the decrease in clean accuracy. By optimizing the agent with a limited amount of clean data, ONP find the most potential backdoor neurons in the second quadrant, despite some mistakenly pruned clean neurons in the fourth quadrant. Although our ONP can be further improved to reduce the number of unnecessarily pruned neurons, it's enough to reveal the significance of optimizing pruning policies for backdoor mitigation.

\subsection{Ablation Studies}
\subsubsection{Impact of Defense Data Size} 
\begin{figure}[t]
\centering
\includegraphics[scale = 0.25]{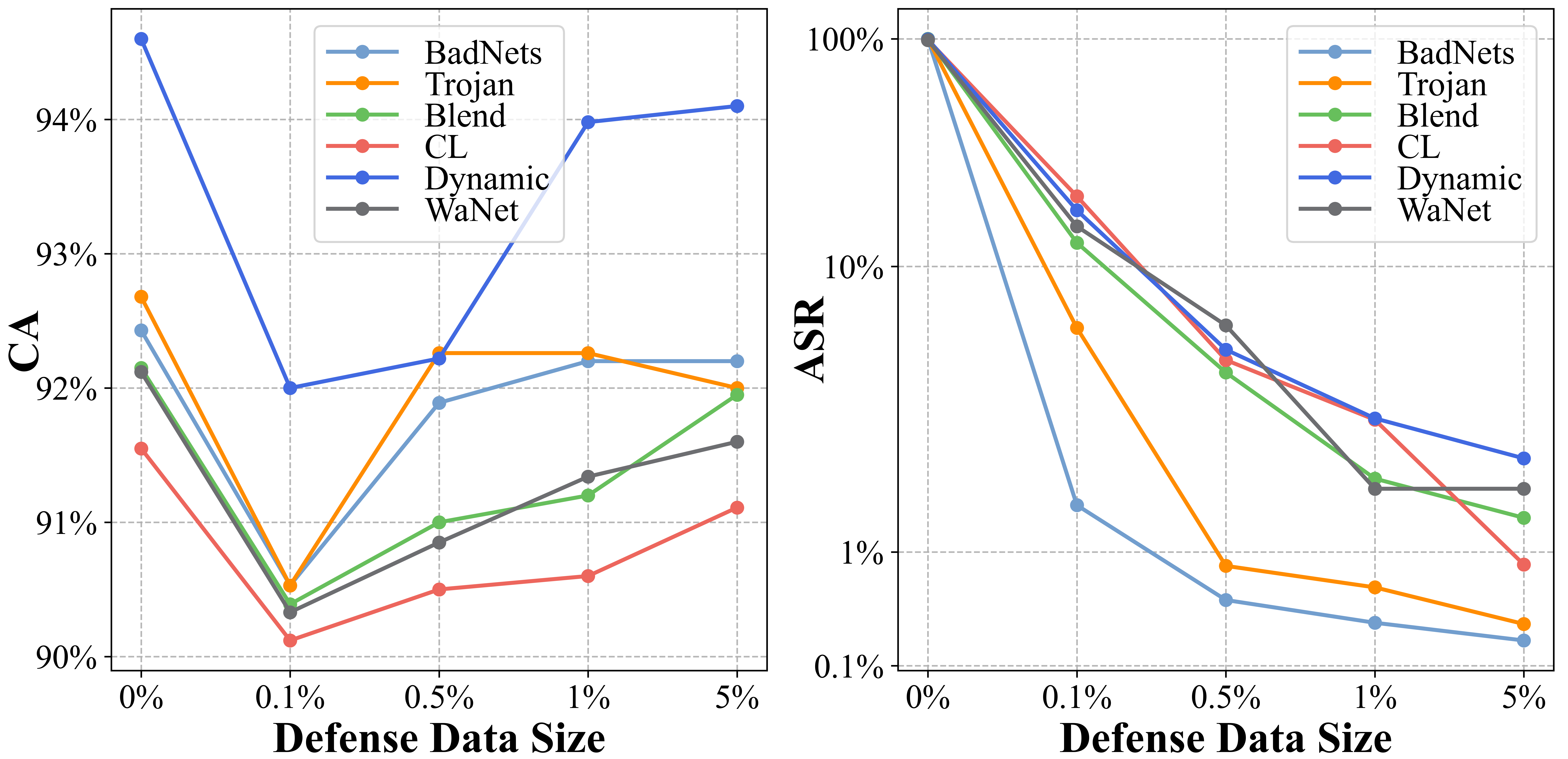}
\caption{Defense performance of ONP with different defense data size}
\label{fig6}
\end{figure}

In this part, we evaluate the impact of defense data size on the optimization of pruning policies. We use 0.1\%(50), 0.5\% (250), 1\% (500) and 5\% (2500) images from the clean CIFAR-10 training set for defense, respectively. Results in Fig \ref{fig6} show that as defense data size increases, the distribution of defense data aligns more closely to the real test set, enhancing the computed reward and further improving CA. Simultaneously, trigger synthesis quality also improves, aiding the RL agent in identifying more backdoor neurons and further reducing ASR. Generally, 1\% defense data is sufficient for ONP to achieve high CA and low ASR against most attacks, making it suitable for data-limited scenarios.

\subsubsection{Choice of Hyperparameters} \label{para}
As is mentioned in Section 3.2, the ONP hyperparameter $\lambda$ balances the backdoor erasing performance and clean accuracy of the pruned model. A higher $\lambda$ makes the agent more conservative and tends to avoid degradation to CA, while a lower $\lambda$ allows for more CA reduction, enabling the agent to prune more backdoor neurons and further reduce ASR. Setting $\lambda = 5$ is recommended for most scenarios.
$\epsilon$ and $\delta$ influence graph construction. $\delta$ filters dormant neurons with small weights to reduce search space for the agent, while $\epsilon$ conserves edges with the highest weight for neurons with similar connection strength. To alleviate the difference between layers and models, we suggest adaptive settings for $\epsilon$ and $\delta$ to conserve 5\% edges and 50\% nodes for each layer, which is proved to be effective for most scenarios.

\subsubsection{Performance Across Architectures and Datasets}
We extend the evaluation of ONP to diverse settings, including VGG16 on MNIST, Pre-Activation ResNet101 on GTSRB, and ResNet34 on YouTubeFace, all attacked by BadNets. ONP is selectively applied to the last 3 convolutional layers or blocks. The defense results are shown in \ref{tab:3}. ONP reduces the ASR down to 0.29\%, 0.36\%, and 0.44\%, respectively, with negligible degradation to CA, proving the robustness of ONP across different layers and architectures.

\subsubsection{Running Time} ONP uses RL to optimize pruning policies and needs computing loss on the defense data to obtain rewards. However, it does not perform back
\begin{wraptable}{r}{0.5\textwidth}
\vspace{-20pt}
\caption{Defense performance of ONP on different architectures and datasets.}

\small
\centering
\begin{adjustbox}{width=0.95\linewidth}
\begin{tabular}{@{}c|cc|cc|cc@{}}
\toprule
\multirow{2}{*}{\begin{tabular}[c]{@{}c@{}}Datasets\end{tabular}} & \multicolumn{2}{c|}{MNIST} & \multicolumn{2}{c|}{GTSRB} & \multicolumn{2}{c}{YouTubeFace} \\ \cmidrule(l){2-7} 
        & CA    & ASR   & CA    & ASR   & CA    & ASR   \\ \midrule
Backdoored  & 98.74 & 100.00 & 96.80 & 100.00 & 97.79 & 100.00 \\ \midrule
ONP         & 98.21 & 0.29 & 96.31 & 0.36   & 97.64  & 0.44\\ \bottomrule
\end{tabular}
\end{adjustbox}

\label{tab:3}
\vspace{-20 pt}
\end{wraptable}
propagation on the backdoored model and only needs to train the PPO agent with fewer parameters. We record the running time of ONP on RTX 3090Ti GPU with 500 CIFAR-10 samples and a ResNet18 model attacked by BadNets. ONP is applied to the last 2 ResNet blocks, and 2 agents are optimized for a total of 200 episodes. It costs ONP 6 minutes and 27 seconds to reduce the ASR to 0.44, slower than the rule-based methods including CLP (1 second), FP (30 seconds) ANP (45 seconds) and RNP (2 minutes and 11 seconds). However, ONP achieves better backdoor mitigation performance, and the computing cost is acceptable compared to retraining the model from scratch (more than 1 hour).

\section{Conclusion}
In this paper, we rethink pruning for backdoor mitigation from an optimization perspective and propose ONP, the first optimization-based pruning framework for backdoor mitigation. We explore the connections between backdoor neurons and convert the DNN into graphs based on the neuron connection strength to expose backdoor neurons. We combine GNN and RL to perform pruning on both the graph and the infected model to search for the optimal pruning policy. Additionally, we investigate the impact of residual connections on backdoor activation and extend pruning strategies for residual connections from model compression to backdoor mitigation. We empirically show the effectiveness of ONP as a backdoor mitigation method and emphasize the significance of optimization for backdoor mitigation. We hope our work can offer some insights for developing more powerful backdoor defense methods in the future.

%
References will then be sorted and formatted in the correct style.
\bibliographystyle{splncs04}
\bibliography{main}
\end{document}